\documentclass[10pt,twocolumn,letterpaper]{article}

\usepackage{iccv}
\usepackage{times}
\usepackage{epsfig}
\usepackage{graphicx}
\usepackage{svg}
\usepackage{amsmath}
\usepackage{amssymb}
\usepackage{adjustbox}
\usepackage{color}
\usepackage{xcolor}
\usepackage{multirow}
\usepackage{booktabs}
\usepackage{bbding}
\usepackage{utfsym}

\newcommand{\bp}[1]{\textcolor{black}{#1}}
\newcommand{\zb}[1]{\textcolor{black}{#1}}
\newcommand{\ct}[1]{\textcolor{black}{#1}}
\newcommand{\phy}[1]{\textcolor{black}{#1}}

\usepackage[breaklinks=true,bookmarks=false]{hyperref}

\iccvfinalcopy 

\def\iccvPaperID{6979} 

\ificcvfinal\fi

\begin{document}

\title{Multi-view Vision-Prompt Fusion Network: Can 2D Pre-trained Model Boost 3D Point Cloud Data-scarce Learning?}


\author{Haoyang Peng$^{1}$, Baopu Li$^{2}$, Bo Zhang$^{3}$, Xin Chen$^{4}$, Tao Chen$^{1}\thanks{ Corresponding author (eetchen@fudan.edu.cn)}~$, Hongyuan Zhu$^{5}$\\
  $^1$Fudan University\quad 
  $^2$Independent Researcher \quad
  $^3$Shanghai AI Laboratory \quad
  $^4$Tencent PCG\quad\\
  $^5$Institute for Infocomm Research(I$^2$R) \& Centre for Frontier AI Research (CFAR), A*STAR, Singapore\vspace{0.2cm}\\
}

\maketitle
\ificcvfinal\thispagestyle{empty}\fi


\begin{abstract}
     Point cloud based 3D deep model has wide applications in many applications such as autonomous driving, house robot, and so on. Inspired by the recent prompt learning in natural language processing, this work proposes a novel Multi-view Vision-Prompt Fusion Network (MvNet) for few-shot 3D point cloud classification. MvNet investigates the possibility of leveraging the off-the-shelf 2D pre-trained models to achieve the few-shot classification, which can alleviate the over-dependence issue of the existing baseline models towards the large-scale annotated 3D point cloud data. Specifically, MvNet first encodes a 3D point cloud into multi-view image features for a number of different views. Then, a novel multi-view prompt fusion module is developed to effectively fuse information from different views to bridge the gap between 3D point cloud data and 2D pre-trained models. A set of 2D image prompts can then be derived to better describe the suitable prior knowledge for a large-scale pre-trained image model for few-shot 3D point cloud classification. Extensive experiments on ModelNet, ScanObjectNN, and ShapeNet datasets demonstrate that MvNet achieves new state-of-the-art performance for 3D few-shot point cloud image classification. The source code of this work will be available soon.
\end{abstract}

\section{Introduction}

Classification of 3D point cloud objects is a critical task in computer vision and has been widely applied in real-world scenarios such as metaverse~\cite{YukeZhu2017TargetdrivenVN}, autonomous driving~\cite{YaodongCui2020DeepLF,MingLiang2018DeepCF,WeijingShi2020PointGNNGN}, etc. Traditional algorithms~\cite{YuZhong2009IntrinsicSS,RaduBogdanRusu2008AligningPC,FedericoTombari2010UniqueSC} design hand-craft rules to represent point clouds, deep learning-based methods~\cite{AlbertoGarciaGarcia2016PointNetA3,CharlesRQi2017PointNetDH,AnhVietPhan2018DGCNNAC,YongbinLiao2023PointCI,SijinChen2023EndtoEnd3D} unify the extraction and learning of point cloud feature using deep neural networks have achieved better performance in various downstream tasks~\cite{YulanGuo2019DeepLF,HangSu2015MultiviewCN,TanYu2018MultiviewHB,YuxingXie2019ARO,GusiTe2018RGCNNRG}.

However, existing deep learning-based approaches often require large amount of point cloud data for a deep neural network to learn and achieve an impressive and stable performance. In 2D computer vision, with large datasets such as ImageNet~\cite{JiaDeng2009ImageNetAL}, COCO~\cite{TsungYiLin2014MicrosoftCC} and PASCAL VOC~\cite{MarkEveringham2010ThePV}, it is possible to pre-train a 2D deep convolutional neural network such as ResNet~\cite{KaimingHe2015DeepRL}, ConvNeXt~\cite{ZhuangLiu2023ACF}, and ViT~\cite{AlexeyDosovitskiy2020AnII}, which can later be fine-tuned for specific downstream tasks, and further to achieve promising and robust performance with only a limited amount of data. On the other hand, existing 3D point cloud methods are usually investigated on much smaller scale datasets, such as ModelNet~\cite{ZhirongWu20143DSA}, ScanobjectNN~\cite{MikaelaAngelinaUy2019RevisitingPC}, and ShapeNet~\cite{AngelXChang2015ShapeNetAI}. Due to the large domain differences between different point cloud datasets and the insufficient amount of data in a single dataset to train a 3D generic feature extractor, learning generic point cloud feature representations using pre-trained 3D deep networks is generally challenging.

\begin{figure}[tb]
\begin{center}
\vspace{6pt}
\includegraphics[width=1.0\linewidth]{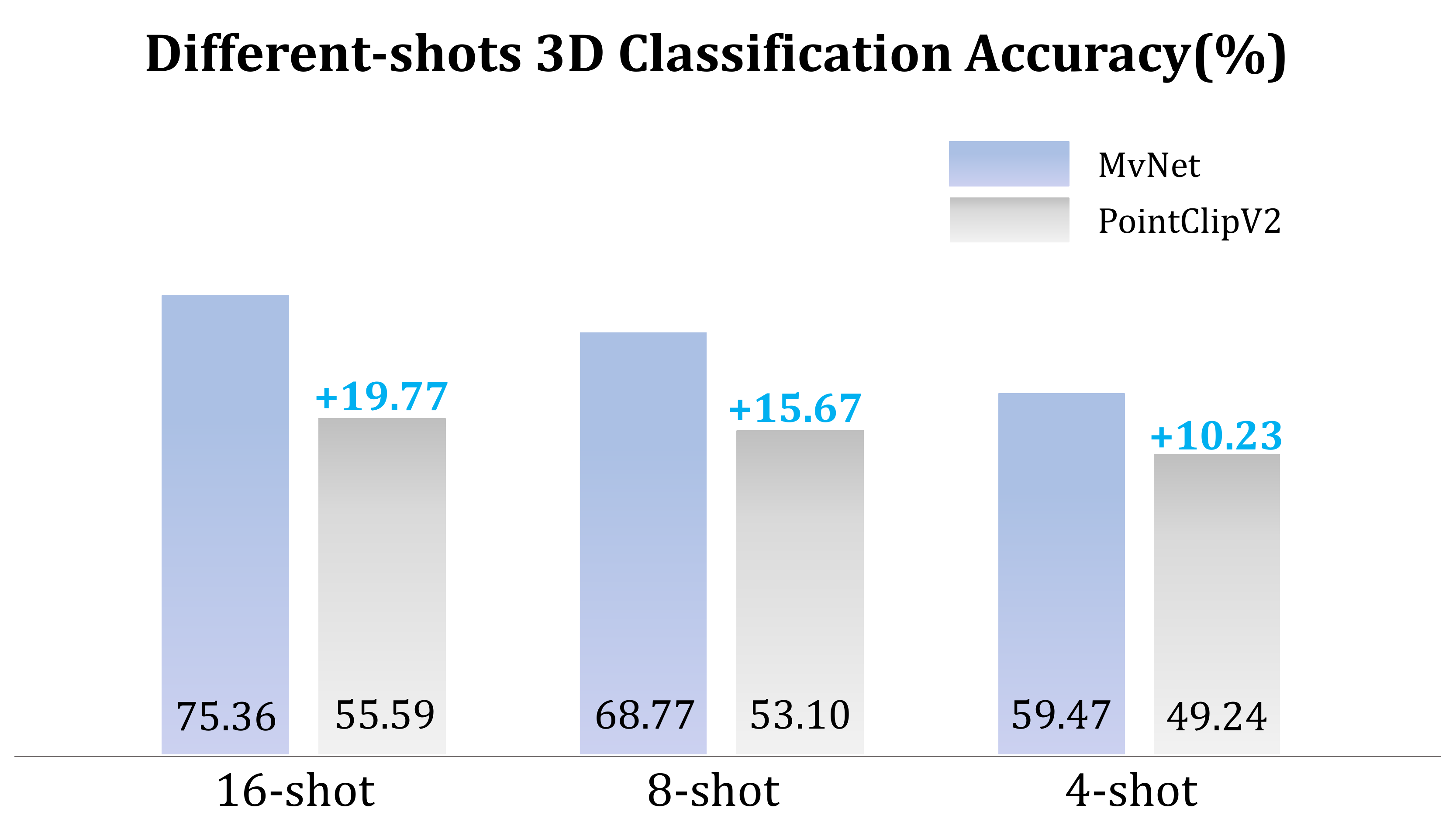}
\end{center}
   \caption{\textbf{Few-shot Performance of MvNet on ScanObjectNN.} Benefit from the rich knowledge of large-scale pre-trained 2D image models, our MvNet significantly outperforms PointCLIPV2 in terms of 3D point cloud classification accuracy under different shots.}
\label{fig: teaser}
\vspace{4pt}
\end{figure}

There are two main types of approaches to addressing the challenge of label scarce issue in 3D point cloud. The first approach is meta-learning-based method~\cite{ye2022makes, ye2023closer} that explores improving model by simulating different few-shot scenarios. The second approach (e.g.,~PointCLIP~\cite{RenruiZhang2023PointCLIPPC}, PointCLIPV2~\cite{XiangyangZhu2022PointCLIPVA}) investigates transferring pre-trained 2D CLIP model to 3D point cloud recognition. The meta-learning-based approach studies a more restrictive few-shot scenario, where the model can see different meta-sets at different stages. In contrast, PointCLIP~\cite{RenruiZhang2023PointCLIPPC} studies the few-shot training without using the meta-sets, which is more practical but challenging. 



In recent years, prompt tuning has emerged in natural language processing (NLP) by adapting pre-trained large-scale language model~\cite{XiangLisaLi2021PrefixTuningOC,BrianLester2021ThePO,XiaoLiu2021GPTUT}, where prompt is inputted into the  pretrained model for the downstream task. Inspired by this, we also exploit to introduce a low-cost prompt for the few-shot 3D point cloud classification task, to address the challenges as above. Furthermore, as mentioned before, the large-scale ImageNet pre-trained 2D model contains rich appearance information of a class of objects, which can be treated as useful prior knowledge for recognizing a 3D object of the same class. This also accords with human visual cognition manner, e.g., we humans can recognize a 3D object in only one-shot, by recalling the previously memorized 2D appearances of the object and matching them with the 3D structure.

\begin{figure*}[tb]
\begin{center}
\vspace{6pt}
\includegraphics[width=1\linewidth]{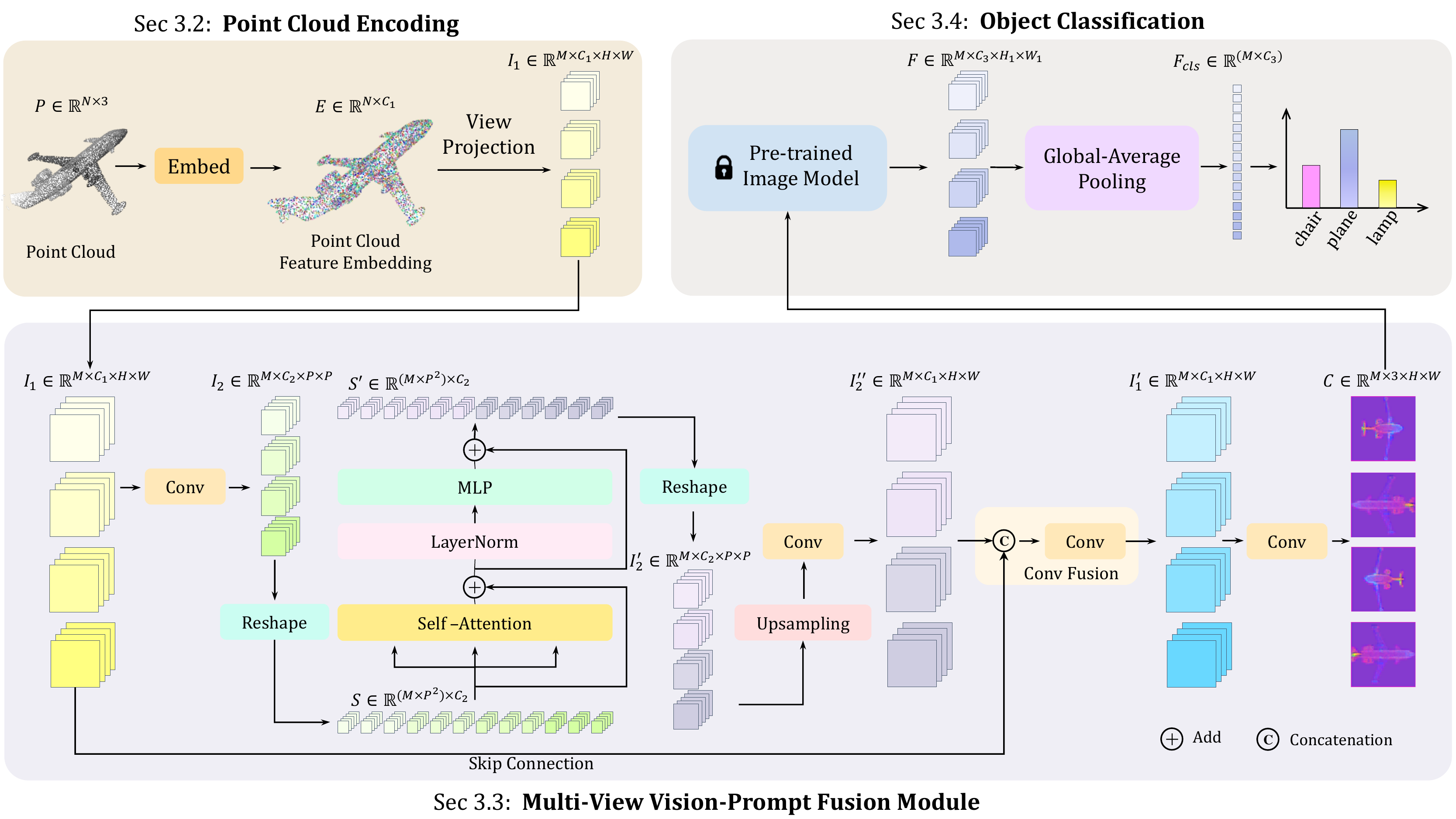}
\end{center}
\vspace{-0.15cm}
   \caption{\textbf{The pipeline of our proposed MvNet.} Using a point cloud $P$ as the input, \phy{we first change its geometric information into a high-dimensional feature space to obtain point cloud feature embedding $E$, and then perform orthogonal view projection to obtain multi-view image features $I_1$}. Then, the $I_1$ are organized into feature sequences $S$ by convolution and sent to the self-attention layer for feature interaction and fusion. Next, the feature sequence $S'$ outputted from the self-attention layer is reshaped and upsampled to the size of the image feature map. We apply a convolution layer to the up-sampled feature map to mitigate the information loss problem during the down-sampling process. The resulting feature map $I''$ is then used as a complement to the original image feature map, and Conv Fusion is performed with it, resulting in a 3-channel multi-view vision prompt $C$ by convolution. \phy{In the object classification phase, the pre-trained image model extracts the features $F$ of $C$ and finally outputs the class score of the point cloud object $P$.}}
\label{fig: pipeline}
\vspace{-0.2cm}
\end{figure*}

Therefore, we consider using low-cost 2D images as prompts for the 3D classification task in this work. Specifically, we advance a Multi-view Vision-Prompt Fusion Network (MvNet) to prompt large-scale pre-trained 2D models, such as ResNet and Vit, to effectively perform a few-shot 3D classification, reducing the dependence of 3D perceptual models on large-scale 3D point cloud data.
We first encode a 3D point cloud to multiple views. Then, we propose a multi-view prompt vision fusion module based on attention mechanism to exchange and fuse information from different views. Finally, we obtain a set of prompts to leverage the prior knowledge of a large-scale 2D pre-trained image model to 3D few-shot learning. As shown in Figure~\ref{fig: teaser}, our MvNet significantly outperforms PointCLIPV2 for 3D point cloud few-shot classification on the ScanObjectNN dataset.

We summarize the contributions of our work as follows: 
\begin{enumerate}
  \item For the first time, we bridge the gap between 2D pre-trained models and 3D point cloud data by prompting 2D pre-trained models with multi-view projected 3D point cloud to achieve 3D point cloud few-shot classification.
  \item We design a new network called MvNet, which firstly encodes multi-view vision prompting on point cloud data, making it possible to use 2D pre-trained models to reason about 3D point clouds. Next, we use the prompting information of multiple views to interact and fuse to produce transferable representations of a point cloud-based object, which can better leverage 2D pre-trained model prior knowledge, thus facilitating effective learning for 3D point clouds.
  \item Extensive experiments validate that the proposed new approach demonstrates new state-of-the-art classification accuracy, outperforming those of PointCLIP and PointCLIPV2.
\end{enumerate}

\section{Related Works}

\noindent \textbf{Visual Pre-trained Model.} In  2D image domain, researchers can use models pre-trained on large-scale image datasets \cite{JiaDeng2009ImageNetAL,TsungYiLin2014MicrosoftCC,MarkEveringham2010ThePV, yuan2023ad, zhang2023uni3d} to fine-tune on specific downstream task datasets by means of transfer learning \cite{AmirRoshanZamir2018TaskonomyDT}, to obtain a more generalizable image model in a cost-effective manner. \bp{The widespread} use of fully supervised models \cite{KarenSimonyan2015VeryDC,AlexeyDosovitskiy2020AnII,PengYe2022StimulativeTO} \bp{such as ResNet~\cite{KaimingHe2015DeepRL}, DenseNet~\cite{GaoHuang2016DenselyCC} and so on has also contributed greatly to the field of computer vision.} At the same time, a number of unsupervised \cite{HangboBao2021BEiTBP,KaimingHe2021MaskedAA,KaimingHe2023MomentumCF} and weakly supervised \cite{AnttiTarvainen2017MeanTA,DavidBerthelot2019MixMatchAH,QizheXie2020SelfTrainingWN} pre-trained models also bring better visual feature representation and adaptability to downstream tasks for the transfer learning process.

\bp{Regarding 3D vision, there are also some impressive} large point cloud feature extraction backbones. For example, PointContrast~\cite{SainingXie2020PointContrastUP} attempts to pre-train point cloud models using unsupervised methods. Point-Bert~\cite{XuminYu2023PointBERTP3} brings the idea of BERT from natural language processing to point clouds. And Point-MAE~\cite{pang2022masked} pre-trains point cloud models using the idea of MAE~\cite{KaimingHe2021MaskedAA}. However, as the amount of data in the training dataset used to pre-train the 3D model is not as rich as the 2D data, a \bp{rather large} domain gap exists between different 3D datasets \cite{ZhirongWu20143DSA,MikaelaAngelinaUy2019RevisitingPC,AngelXChang2015ShapeNetAI,AngelaDai2017ScanNetR3}.
In comparison, 2D pre-training models are more capable in visual representation, \bp{and there has been a lot of wonderful 2D pre-trained models that are off-the-shelf}. Our work uses large-scale 2D pre-trained models for \bp{few sample} 3D point cloud classification tasks, effectively transferring knowledge from the 2D image domain to the 3D point cloud domain.
\vspace{6pt}

\noindent \textbf{Prompt Learning.} Previous transfer learning~\cite{AmirRoshanZamir2018TaskonomyDT} methods based on fine-tuning \bp{may suffer from different problems such as {a rather large gap} between targets of downstream tasks,}  the pre-trained target being too large and the fine-tuning process relying on a large amount of label data.
To \bp{mitigate such problems}, a new fine-tuning paradigm based on pre-trained language models, Prompt-Tuning, \bp{has been recently} proposed.
\bp{Such a scheme aims to improve the performance via prompts, and it allows} language models to achieve \bp{encouraging} results in few-shot or zero-shot scenarios. Subsequently, continuous prompting methods \cite{XiangLisaLi2021PrefixTuningOC,XiaoLiu2021GPTUT,XiaoLiu2023PTuningVP}, which use fewer parameters to obtain better performance, have become the main choice.

In recent years, some works have attempted to introduce the idea of prompt tuning into the field of vision. VPT~\cite{MenglinJia2022VisualPT} uses Visual Prompt Tuning for the first time to transfer learning from large pre-trained visual models. P2P~\cite{wang2022p2p} takes the lead in point cloud classification and part segmentation tasks using prompts generated by point cloud projections. In our work, we generate visual prompts that make better use of the knowledge of large-scale pre-trained models through a process of fusing multi-view features of the point cloud.
\vspace{6pt}

\noindent \textbf{Few-shot 3D Point Cloud Classification.} 
\bp{Few shot learning has been a popular research topic~\cite{yuan2023bi3d, RenruiZhang2023PointCLIPPC, XiangyangZhu2022PointCLIPVA, AlecRadford2021LearningTV} in recent years due to its impressive ability in handling the real application screnario.} 
A \bp{typical approach to few shot learning} is to use a \ct{fixed} few-shot dataset throughout the training process and generate a model that is more generalizable. The PointCLIP family \cite{RenruiZhang2023PointCLIPPC,XiangyangZhu2022PointCLIPVA} has already explored along this direction. PointCLIP~\cite{RenruiZhang2023PointCLIPPC} introduces CLIP~\cite{AlecRadford2021LearningTV}, a pre-trained multimodal using large-scale 2D image text pairs, to the 3D few-shot classification problem for the first time. This method converts 3D point clouds into 2D images of multiple views by direct multi-view depth map projection of the point clouds, and then extracts the visual features of each view using the pre-trained visual coder in CLIP~\cite{AlecRadford2021LearningTV}. It obtains the final category scores by fusing them with natural language description features of multiple categories, thus achieving outstanding few-shot classification performance. \bp{Later,} PointCLIPV2~\cite{XiangyangZhu2022PointCLIPVA} improves the point clouds projection method in PointCLIP and introduces more statements to describe the target species, significantly improving the model's few-shot and zero-shot classification performance. However, the disadvantage of these two methods is the excessive amount of parameters. 
In this work, we use a fixed few-shot training set to train a model using only image modalities, ultimately achieving better performance with fewer parameters. 

\section{Method}

\subsection{Overview}

The overall framework of our Multi-view Vision-Prompt Fusion Network (MvNet) is illustrated in Figure~\ref{fig: pipeline}. The network architecture consists of \bp{three} components: 1) a lightweight 3D point cloud encoder that encodes point cloud data from 3D coordinates into 2D multi-channel images; 2) a  
Multi-view Vision Prompt 
module that takes the 2D multi-view features generated by the point cloud encoder, and interacts the features between views, generates complementary features for that view based on features from other views, and further fuses them with the original features to project a three-channel vision prompt 
\bp{that works well with the 2D image pre-trained model}; \bp{please be noted that here the 2D image pre-trained model refers to some typical networks such as ResNet or ViT.} 3) an object classification module that uses a large 2D pre-trained image model to extract features from the multi-view vision prompt, and then outputs a class probability vector for the point cloud object based on the extracted multi-view features using a simple classification head.

Our proposed network can be trained in an end-to-end manner. In the forward process, inspired by ~\cite{wang2022p2p}, we first encode the point cloud into multi-view image features and then fuse the multi-view features to get the complementary information of the original features and merge them with the original parts. The image features containing other view information are projected into three-channel multi-view vision prompts, and the prompts obtained in this way can better fit the knowledge of the 2D image pre-training model. During the back-propagation phase, the frozen pre-trained image model can guide the Multi-view Vision Prompt module to generate vision prompts that better fit the image model.



\subsection{Point Cloud Encoding}

In ~\cite{RenruiZhang2023PointCLIPPC}, the 2D depth map of the corresponding view is obtained directly by a perspective projection of the point cloud. Due to the sparse nature of the point cloud, the projection of the point cloud results in distinctive dot-like patterns in the image, which will cause a significant hindrance to the recognition and extraction of image features by 2D image pre-training models. Inspired by~\cite{wang2022p2p}, we wish to use Geometry-Preserved Projection to project the point cloud from 3D coordinates into 2D image form while projecting the geometric texture information of the point cloud into the high-dimensional feature space for subsequent multi-view feature fusion.




\phy{In detail, given an input point cloud  $P\in \mathbb{R}^{N\times 3}$  with  $N$  points, we first use a lightweight convolution to change its geometric coordinate information into a high-dimensional feature space to obtain the high-dimension point features $P'\in \mathbb{R}^{N\times C_1}$ of the point cloud. Inspired by~\cite{CharlesRQi2017PointNetDH},  we then find k-nearest neighbor points $p_j$ of each point $p_i$ in the point cloud $P$ and obtain their feature embedding $e_i$ and $e_j$, to learn the point cloud local geometry information by optimizing the point set feature embedding $E_i$ formulated as below:}

\begin{equation}
    E_i = \mathrm{maxpool}(\phi (\mathrm{concat}(e_i, e_j - e_i)))
\vspace{6pt}
\end{equation}

\noindent where $\phi$ is a lightweight convolutional neural network used to fuse point set feature, and point set feature $E_i \in \mathbb{R}^{1 \times C_1} \subseteq E \in \mathbb{R}^{N \times C_1}$.


\phy{Just like \cite{wang2022p2p}, in order to project a point cloud object into uniformly-sized $M$-view 2D image features, we first calculate the length, width, and height of the point cloud object and then calculate which grid in the $x-y$ plane each point $(x,y,z)$ of the object belongs to from $M$ different perspectives. Then, the point cloud features belonging to the same grid are summed as features of that pixel according to the correspondence between the point cloud coordinates and the 2D grid coordinates.}



Finally, we can get the multi-view image features $I_1 \in  \mathbb{R}^{M\times C_1\times H \times W}$ projected from the geometric features of the point cloud in different views, where $C_1$, $H$, and $W$ are the channel, height, and width of the multi-view image features, respectively.  In order to obtain a coordinate distribution in line with the real-world image of the object (dense and uniform), we shift the multi-view image features so that features can be propagated to pixel points that do not have a point cloud feature set due to point cloud sparsity.

\vspace*{8pt}

\subsection{Multi-View Vision-Prompt Fusion}
 
\phy{After obtaining the point cloud multi-view projection image features $I_1$ from the point cloud encoding output, we wish the information between views to be used by other views will supplement their own information, laying the groundwork for obtaining a comprehensive and complementary set of multi-view visual prompting later on}. For this purpose, we propose the Multi-view Vision Prompt Module to help interact image features from a single view with \bp{those} from all views.

\vspace*{8pt}
\noindent \textbf{Multi-view Feature Sequence Generation.}  Specifically, \zb{a}fter obtaining multi-view image features $I_1 \in  \mathbb{R}^{M\times C_1\times H \times W}$  of the input point cloud,  we can get a set of patch token feature maps $I_2 \in \mathbb{R}^{M \times C_2 \times P \times P}$ by convolution operations as: 



\begin{equation} \label{eq1}
    I_2 = I_1 * k_1
  \vspace{6pt}
\end{equation}

\noindent where $k_1 \in \mathbb{R}^{C_2\times C_1\times 7\times 7}$ denotes the convolution kernel \bp{with a size of} $7$, $*$ is the convolution operator,  $C_2$ is the dimension of the feature map after convolution, 
and $(P, P)$  is the resolution of the image feature map after convolution. Then we reshape the image feature map to get a multi-view image feature sequence $S \in \mathbb{R}^{(M\times P^2)\times C_2}$, which will be fed to the self-attention layer for interaction and fusion of the multi-view features. 

\vspace*{8pt}
\noindent \textbf{Multi-view Feature Interaction.} After obtaining the multi-view feature sequence $S$, we first normalize it to a triplet of ${\rm Q}\in \mathbb{R}^{(M\times P^2)\times C_2}$, ${\rm K}\in \mathbb{R}^{(M\times P^2)\times C_2}$ and ${\rm V}\in \mathbb{R}^{(M\times P^2)\times C_2}$ as the input token sequence of the Attention Fusion module, where $(K \times P^2)$ is the length of the input token sequence, and $C_2$ is the feature dimension of each token. Then we use the self-attention~\cite{AshishVaswani2017AttentionIA} to compute the attention values between queries and keys. Every output token is a weighted sum of all tokens using the attention values as weights, formulated as:

\begin{equation} \label{eq1}
    {\rm Attention(\textbf{Q}, \textbf{K}, \textbf{V})} = {\rm softmax(\textbf{Q}\textbf{K}^T / \sqrt{C_2})\textbf{V}}
  \vspace{6pt}
\end{equation}

For a point cloud object, we can get its comprehensive view via different components corresponding to multi-view projections of the point cloud. 
\phy{In other words, when the multi-view image feature sequence is applied by the self-attention layer, the features in a region of one of the views can be enriched by the long-range dependencies between all the point cloud projection region features of all the views, thus endowing the model a more comprehensive understanding of the region.} Then, by extracting the regional correlation between views, we can effectively enhance the more important parts of this view with information from other views for the subsequent vision prompt.
 
Next, the image feature sequences $S' \in \mathbb{R}^{(M\times P^2)\times C_2}$ obtained through feature interaction and fusion are reorganized into patch token feature maps $I_2' \in \mathbb{R}^{M\times C_2\times P \times P}$. In order to fuse the resulting patch token feature map $I_2'$ with the original feature map $I_1$, we interpolate and upsample $I_2'$ to the same \bp{size} as $I_1$, and restore as much complementary information as possible by convolution with a kernel size of $3*3$. The final convolved image feature map is $I_2'' \in \mathbb{R}^{M\times C_1\times H \times W}$.

\begin{table*}[tb]
\centering
\vspace{6pt}
\begin{adjustbox}{width=1.0\linewidth}
\begin{tabular}{cccccccccc}
\toprule
Method              &           & MN40-4-shot & MN40-8-shot & MN40-16-shot & SN-4-shot & SN-8-shot & SN-16-shot & Trainable Params & All Params \\
\midrule
PointNet ~\cite{AlbertoGarciaGarcia2016PointNetA3} &  & 54.74 & 63.66 & 72.21 & 26.47 & 34.97 & 35.84 & 13.24MB    & 13.24MB    \\
PointNet++ ~\cite{CharlesRQi2017PointNetDH}        &  & 72.44 & 78.00 & 79.39 & 40.70 & 47.68 & 54.98 & 5.62MB     & 5.62MB     \\
CurveNet ~\cite{AAMMuzahid2021CurveNetCM}          &  & 69.56 & 75.57 & 80.77 & 26.12 & 30.59 & 35.22 & 8.16MB     & 8.16MB     \\
SimpleView ~\cite{AnkitGoyal2021RevisitingPC}      &  & 57.98 & 68.74 & 78.69 & 29.22 & 32.40 & 37.38 & 1067.64MB  & 1067.64MB  \\
PointCLIP ~\cite{RenruiZhang2023PointCLIPPC}       &  & 77.07 & 81.35 & 87.20 & 46.14 & 50.00 & 55.50 & 84.06MB    & 473.18MB   \\
PointCLIPV2 ~\cite{XiangyangZhu2022PointCLIPVA}    &  & 78.92 & 84.60 & 89.55 & 49.24 & 53.10 & 55.59 & \textgreater{}84.06MB & \textgreater{}473.18MB  \\ 
\midrule
\textbf{Our Method(RN-18)} &  & 82.37 & 85.81 & 89.25 & 49.65 & 60.96 & 70.09 & 5.90MB   & 49.33 MB   \\ 
\textbf{Our Method(RN-50)} &  & 82.61 & \textbf{86.66} & 90.31 & 54.51 & 65.37 & 65.89 & 5.90MB   & 103.22MB   \\ 
\textbf{Our Method(CN-L)} &  & \textbf{84.88} & 86.10 & \textbf{91.16} & \textbf{59.47} & \textbf{68.77} & \textbf{75.36} & 5.90MB  & 759.23MB   \\
\bottomrule
\end{tabular}
\end{adjustbox}
\vspace{-0.1cm}
\caption{\textbf{Few-shot 3D classification results on ModelNet~\cite{ZhirongWu20143DSA} and ScanObjectNN~\cite{MikaelaAngelinaUy2019RevisitingPC}.} We adopt the PB\_T50\_RS split of ScanObjectNN for a fair comparison. With prior knowledge from the ImageNet-pre-trained 2D model, the 3D classification accuracy on ScanObjectNN sampled from the real world can be significantly improved. RN18 and RN50 denote the ResNet-18 and ResNet-50 backbones~\cite{KaimingHe2015DeepRL}, respectively. CN-L denotes ConvNeXt-Large~\cite{ZhuangLiu2023ACF} backbone. MN40 and SN indicate the ModelNet~\cite{ZhirongWu20143DSA} and ScanObjectNN~\cite{MikaelaAngelinaUy2019RevisitingPC}, respectively.}
\label{tab:mian_result}
\vspace{-6pt}
\end{table*}
\vspace*{8pt}

During the feature fusion phase, we fuse the feature map output from the upsampling layer with the original image feature map as additional information. Specifically, we achieve this by concatenating the output feature map with the original feature map along the dimension of the channel, \ct{going through} 2D convolution with a kernel of $1*1$ followed by a GELU~\cite{DanHendrycks2016GaussianEL} activation function, which is used to fuse the pixel features for each feature map. With the Multi-view Feature Fusion Module, we can get a more informative and expressive set of multi-view features $I_1'$ that have been interacted and fused with multiple views, formulated as: 

\begin{equation} \label{eq2}
    I_1'=(I_1 \oplus I_2'')*k_2
\vspace{6pt}
\end{equation}

\noindent where $\oplus$ represents concatenation along the channel dimension and $k_2 \in \mathbb{R}^{C_1\times (2\times C_1)\times 1\times 1}$ denotes the convolution kernel, and the stride of convolution is $1$.

\vspace*{8pt}
\noindent \textbf{Multi-view Vision Prompt Generation.} A simple, lightweight neural network can then be used to project the expressive image features $I_1'$ into multi-view vision prompts $C\in \mathbb{R}^{M\times 3\times H\times W}$ that better fit the pre-trained 2D image model. Again, this network consists of a convolutional layer followed by a GELU activation function, but with a convolutional kernel size of $3*3$, for better aggregation of local image features and image inflation. This gives 2D visual prompting the ability to better match with 2D pre-trained image models.

\phy{Thus, we have a set of information-rich multi-view vision prompts with great adaptability. During the forward propagation of the model in the training phase, the Multi-view Vision Prompt module tries to provide a set of prompts with the same shape and size as the real image based on the multi-view projection features of the 3D point cloud and the large-scale 2D pre-trained image model extracts the features of this `images'.
In the back-propagation process, the back-propagated gradients can be seen as the pre-trained image model pointing out the generic feature embedding information for the vision prompts based on what it has learned before. 
Due to the rich knowledge of multiple views of objects in the large-scale pre-trained 2D image model, and the ability of the Multi-view Vision Prompt Module to generate vision prompts for multiple views simultaneously, 
the Multi-view Vision Prompt Module is highly adaptive and capable of learning. The Multi-view Vision Prompt Module can thus effectively bridge the knowledge gap between 3D point cloud objects and 2D modalities to comprehensively understand point cloud objects in a few-shot context.}

\vspace*{8pt}

\subsection{Object Classification}
 \label{Object Classification}
 
\noindent \textbf{Feature Extraction.}  After obtaining multi-view object representations, we apply a large-scale 2D image backbone pre-trained on ImageNet~\cite{JiaDeng2009ImageNetAL} to perform feature extraction. By feeding a set of 2D multi-view vision prompts $C\in \mathbb{R}^{M\times 3\times H\times W}$ generated by the Multi-view Vision Prompt module to the pre-trained 2D image model, we can obtain the multi-view features $F \in \mathbb{R}^{M\times C_3 \times H_1 \times W_1}$. Then we use these rich multi-view features for the later object classification.


\vspace*{8pt}
\noindent \textbf{Classification via Multi-view Features. }  For 3D point cloud object classification, we utilize high-dimensional multi-view features extracted from the 2D image pre-trained model and a simple structured multi-view feature classification head.

In particular, given a set of multi-view image features $F \in \mathbb{R}^{M\times C_3 \times H_1 \times W_1}$,  we \bp{take} a 2D adaptive average pooling layer for feature dimension reduction. The resulting features are then flattened into a one-dimensional tensor $F_{cls} \in \mathbb{R}^{(M\times C_3)}$, and sent to a fully connected layer to obtain the class score of the point cloud object: $p = {\rm softmax}(F_{cls})$. The cross-entropy loss between the predicted category scores and the ground-truth labels for the input object is used as the final optimization goal.


\section{Experiments}

\vspace*{8pt}
\subsection{Dataset and Implementation Details}
\label{ImplementationDetails}

\noindent \textbf{Datasets.}  We perform point clouds few-shot classification experiments on \textbf{ModelNet} \cite{ZhirongWu20143DSA}, \textbf{ScanObjectNN} \cite{MikaelaAngelinaUy2019RevisitingPC} and \textbf{ShapeNet} \cite{AngelXChang2015ShapeNetAI}. \textbf{ModelNet} contains 40 CAD models and is a standard dataset for evaluating point cloud deep learning models for semantic segmentation, instance segmentation, and classification. \textbf{ScanObjectNN} is a realistic 3D point cloud classification dataset that is challenging due to its background, missing parts, and deformation characteristics. It comprises occluded objects extracted from real-world indoor scans and contains 2,902 3D objects from 15 categories. We take the PB\_T50\_RS split of ScanObjectNN for comparison. \textbf{ShapeNet} is a dataset of 16 classes selected from the ShapeNetCore dataset and annotated with semantic information, comprising 16,846 samples, which we use here to perform few-shot point cloud classification experiments to validate the effectiveness of our model. 

\vspace*{8pt}

\noindent \textbf{Implementation Details.}  During our experiments, we freeze all parameters of the large-scale pre-trained 2D image model, except for the batch normalization~\cite{SergeyIoffe2015BatchNA} layer, and replace the classification head with our proposed multi-view feature classification head as \bp{discussed} in Subsection~\ref{Object Classification}. Regarding the specific operation of our model, the number of nearest neighbors is $k = 32$, and $\phi$ represents a convolutional neural network with GELU as the activation function. Then we utilize AdamW~\cite{IlyaLoshchilov2017DecoupledWD} optimizer and CosineAnnealing~\cite{IlyaLoshchilov2016SGDRSG} scheduler, with  a learning rate of $5e^{-4}$ and weight decay of $5e^{-2}$ to optimize our model. \bp{All our experiments are carried out with PyTorch on a GPU machine of NVidia Geforce RTX 3090.} We train our model with a batch size of 16 for 300 epochs. During training, we perform random rotations of the point cloud by means of a rotation matrix for training data augmentation, and the rotation angle $\alpha$, $\beta$ are randomly chosen from $[-\pi, \pi]$ and $[-0.4\pi,-0.2\pi]$. To better evaluate model performance during the testing phase, we use test time augmentation~\cite{GuotaiWang2019AleatoricUE} to evaluate the test set point clouds by randomly rotating them several times and finally use the result with the most votes as the classification result for that point cloud.

\begin{table}[t]
\begin{adjustbox}{width=1.0\linewidth}
\begin{tabular}{ccccccccc}
\toprule
\multirow{2.5}{*}{Datasets}     & \multirow{2.5}{*}{Methods} & \multirow{2.5}{*}{Backbone} &  & \multicolumn{5}{c}{Different shot(s)} \\ \cmidrule{5-9} 
                              &          &        &  & 1     & 2     & 4     & 8     & 16    \\ 
                              \midrule
\multirow{4}{*}{ModelNet40}   & Baseline & RN-18  &  & 43.92 & 63.77 & 71.88 & 76.01 & 84.60 \\
                              & Baseline & RN-50  &  & 45.62 & 65.47 & 73.98 & 77.99 & 85.21 \\
                              & MvNet    & RN-18  &  & 66.39 & 74.18 & 82.37 & 85.81 & 88.20 \\
                              & MvNet    & RN-50  &  & 67.17 & 77.95 & 82.61 & 86.66 & 88.45 \\ 
                              & MvNet    & CN-L   &  & 67.74 & 77.22 & 84.88 & 86.10 & 89.82 \\
                              \midrule
\multirow{4}{*}{ScanObjectNN} & Baseline & RN-18  &  & 27.89 & 34.07 & 41.91 & 42.50 & 55.86 \\
                              & Baseline & RN-50  &  & 25.88 & 30.13 & 41.84 & 45.31 & 55.96 \\
                              & MvNet    & RN-18  &  & 35.63 & 37.50 & 49.65 & 60.96 & 70.00 \\
                              & MvNet    & RN-50  &  & 37.64 & 47.46 & 54.51 & 65.37 & 65.89 \\ 
                              & MvNet    & CN-L   &  & 38.48 & 54.58 & 59.47 & 68.77 & 75.36 \\
                              \midrule
\multirow{4}{*}{ShapeNet}     & Baseline & RN-18  &  & 47.45 & 70.04 & 81.00 & 89.80 & 93.07 \\
                              & Baseline & RN-50  &  & 44.22 & 67.91 & 82.88 & 90.01 & 93.49 \\
                              & MvNet    & RN-18  &  & 74.49 & 77.76 & 92.69 & 93.00 & 95.78 \\
                              & MvNet    & RN-50  &  & 64.30 & 89.45 & 87.54 & 94.53 & 96.03 \\ 
                              & MvNet    & CN-L   &  & 75.78 & 84.82 & 86.70 & 95.89 & 95.75 \\
                              \bottomrule
\end{tabular}
\end{adjustbox}
\vspace{-0.1cm}
\caption{\phy{OAcc($\%$) of experiments using different 2D pre-trained backbones under different shots, where all shots use the 4-view projection.}}
\label{tab:different_shots}
\vspace{-0.25cm}
\end{table}

\begin{table*}[tb]
\begin{adjustbox}{width=1.0\linewidth}
\begin{tabular}{cccclcccclcccc}
\toprule
\multicolumn{4}{c}{ResNet-18}                                                                                                         &  & \multicolumn{4}{c}{ResNet-50}                                                                                                         &  & \multicolumn{4}{c}{ConvNeXt-Large}                                                                                                    \\ \cmidrule{1-4} \cmidrule{6-9} \cmidrule{11-14} 
Attention Fusion & Conv Fusion & \# views & oAcc  &  & Attention Fusion & Conv Fusion & \# views & oAcc  &  & Attention Fusion & Conv Fusion & \# views & oAcc  \\ 
\cmidrule{1-4} \cmidrule{6-9} \cmidrule{11-14} 
\scalebox{0.75}{$\usym{2613}$}  & \scalebox{0.75}{$\usym{2613}$} & 1 & 84.60 &  & \scalebox{0.75}{$\usym{2613}$} & \scalebox{0.75}{$\usym{2613}$} & 1 & 85.21 & & \scalebox{0.75}{$\usym{2613}$} & \scalebox{0.75}{$\usym{2613}$} & 1 & 86.34      \\
\scalebox{0.75}{$\usym{2613}$}  & \scalebox{0.75}{$\usym{2613}$} & 4 & 86.10 &  & \scalebox{0.75}{$\usym{2613}$} & \scalebox{0.75}{$\usym{2613}$} & 4 & 87.76 & & \scalebox{0.75}{$\usym{2613}$} & \scalebox{0.75}{$\usym{2613}$} & 4 & 86.42      \\
\checkmark & \scalebox{0.75}{$\usym{2613}$} & 4 & 87.84 &  & \checkmark & \scalebox{0.75}{$\usym{2613}$} & 4 & 88.12 &  & \checkmark & \scalebox{0.75}{$\usym{2613}$} & 4 & 89.70      \\
\checkmark & \checkmark & 2 & 87.31 &  & \checkmark & \checkmark & 2 & 85.73 &  & \checkmark & \checkmark & 2 & 87.31 \\
\checkmark & \checkmark & 4 & 88.20 &  & \checkmark & \checkmark & 4 & 88.45 &  & \checkmark & \checkmark & 4 & 89.82 \\
\checkmark & \checkmark & 6 & 89.05 &  & \checkmark & \checkmark & 6 & 89.22 &  & \checkmark & \checkmark & 6 & 91.08 \\
\checkmark & \checkmark & 8 & 89.25 &  & \checkmark & \checkmark & 8 & 90.31 &  & \checkmark & \checkmark & 8 & 91.16 \\ 
\bottomrule
\end{tabular}
\end{adjustbox}
\vspace{-0.1cm}
\caption{Ablation study on changing the number of projection views: 16-shot experiment results of MvNet based on \textbf{ResNet-18}, \textbf{ResNet-50}, and \textbf{ConvNeXt-Large} backbone on ModelNet.}
\label{tab:ablation_study}
\vspace{-0.3cm}
\end{table*}

\vspace*{8pt}
\subsection{Experimental Settings and Evaluation Metric}
\label{Experimental Settings}
\noindent \textbf{Experimental Settings.} \phy{We take two of the most classical (ResNet-18 and ResNet-50~\cite{KaimingHe2015DeepRL}), a recent (ConvNeXt-Large~\cite{ZhuangLiu2023ACF}) large-scale 2D pre-trained image models as the backbone of MvNet for adequate experiments.} We conduct experiments on three public benchmark datasets under 1, 2, 4, 8, and 16 shots. For the $K$-shot experiment setting, we randomly sample $K$ point clouds of each category from the training set. For the selection of the number of projected views, we conduct experiments under 2, 4, 6, and 8 views, \phy{specific details on views can be found in the appendix}. \bp{Besides,} we use the point cloud projection visual prompt of a single view as the baseline for our experiments.

\noindent \textbf{Evaluation Metric.} \bp{We use overall accuracy(oAcc)  to evaluate the performance of the model. Overall accuracy is the proportion of all samples correctly classified by the model, which is calculated via dividing the number of samples correctly classified with the model by the total number of samples. }


\vspace*{8pt}
\subsection{Main Results}

In Table~\ref{tab:mian_result}, we show the results of the few-shot classification of MvNet and compare them with several recent representative methods: PointNet~\cite{AlbertoGarciaGarcia2016PointNetA3}, PointNet++~\cite{CharlesRQi2017PointNetDH}, CurveNet~\cite{AAMMuzahid2021CurveNetCM}, SimpleView~\cite{AnkitGoyal2021RevisitingPC}, PointCLIP~\cite{RenruiZhang2023PointCLIPPC}, and PointCLIPV2~\cite{XiangyangZhu2022PointCLIPVA}. As shown, MvNet outperforms previous methods on both ModelNet and ScanObjectNN. \bp{More} importantly, MvNet significantly improves few-shot classification performance on ScanObjectNN. Compared with PointCLIPV2, MvNet improves 16-shot accuracy by $\textbf{+1.8\%}$ on \textbf{ModelNet} and $\textbf{+35.5\%}$ on \textbf{ScanObjectNN}. In addition, MvNet achieves a 16-shot accuracy of $\textbf{91.16\%}$ on \textbf{ModelNet} and $\textbf{75.36\%}$ on \textbf{ScanObjectNN}, which achieves \bp{new state-of-the-art performance} in known methods.
In Table~\ref{tab:different_shots}, we show the performance of MvNet(4-view) and Baseline with different backbones on ModelNet, ScanObjectNN, and ShapeNet with 1, 2, 4, 8, and 16 shots respectively. The 4-View MvNet significantly outperforms the single-view baseline without multi-view feature fusion for each of the few-shot experimental settings. In the case of MvNet, when the number of views is fixed, the greater the ability of the large-scale pre-trained image model to extract features, the better the model performs in terms of few-shot classification.

\begin{figure}[t]
\begin{center}
\includegraphics[width=0.75\linewidth]{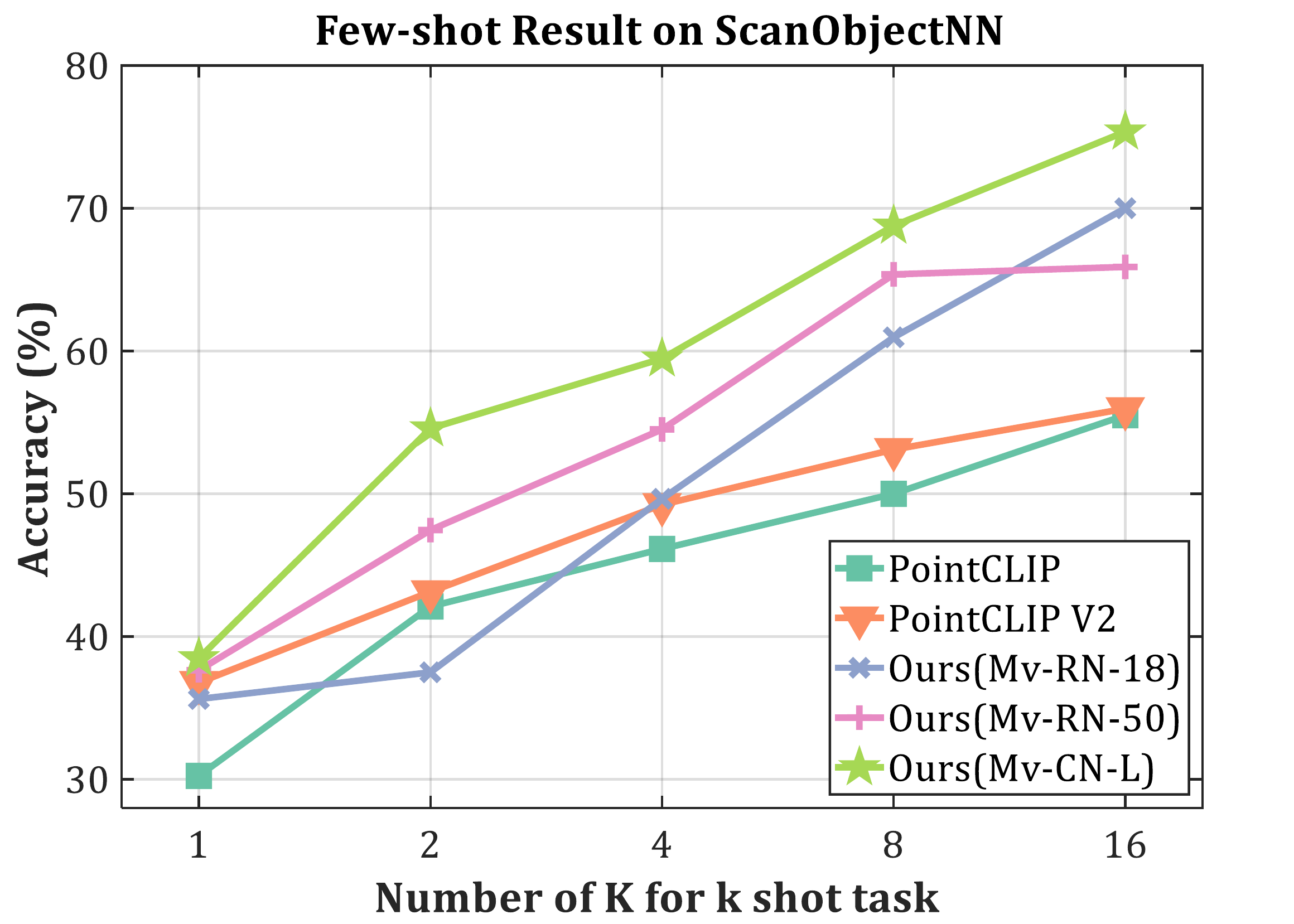}
\end{center}
\vspace{-0.2cm}
   \caption{Few-shot 3D Classification on \textbf{ScanObjectNN}~\cite{MikaelaAngelinaUy2019RevisitingPC}. We adopt the PB T50 RS split of ScanObjectNN for a fair comparison.}
\label{fig: few-shot-scanobjectnn}
 \vspace{-4pt}
\end{figure}

As shown in the Figure~\ref{fig: few-shot-scanobjectnn}, our MvNet significantly outperforms the PointCLIP family for several different few-shot classifications on ScanObjectNN when using powerful backbones such as ResNet-50 and ConvNeXt-Large. At the same time, an average-performing backbone such as ResNet-18 can also achieve promising results in 8-shot and 16-shot experiments. 

\begin{figure}[t]
\begin{center}
\includegraphics[width=0.75\linewidth]{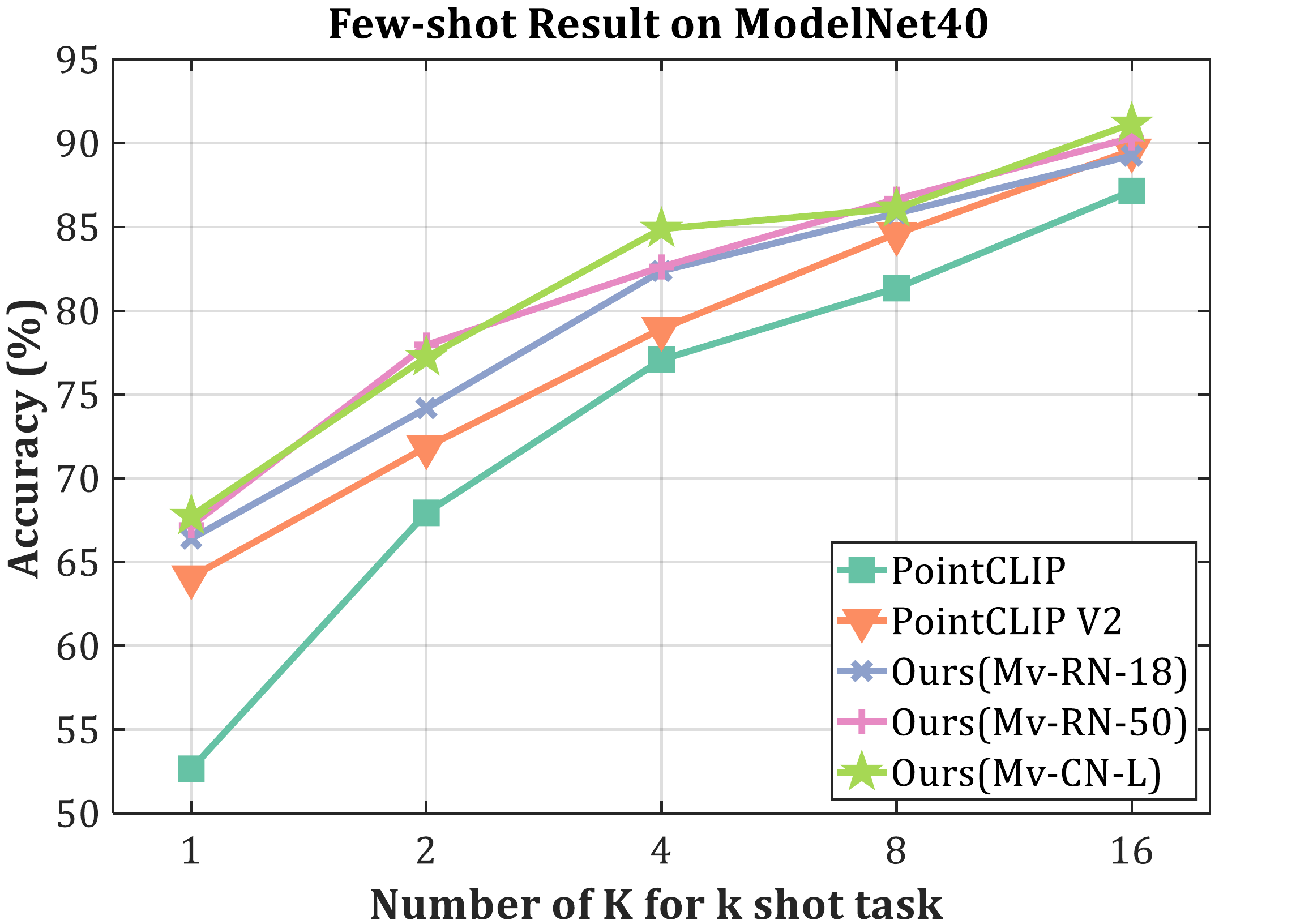}
\end{center}
\vspace{-0.2cm}
   \caption{Few-shot 3D Classification on \textbf{ModelNet40}~\cite{ZhirongWu20143DSA}.}
\label{fig: few-shot-modelnet}
\vspace{-12pt}
\end{figure}

\phy{As illustrated in the Figure~\ref{fig: few-shot-modelnet}, the classification performance of our various backbones based MvNet on ModelNet40 also shows a slight improvement compared to other methods. It is worth noting that, as shown in the Table~\ref{tab:mian_result}, our method can achieve comparable or even better performance with a much smaller number of all parameters and trainable parameters than PointCLIP and PointCLIPV2 on ModelNet40.}

From the experimental results, it is clear that our method can effectively bridge 3D point cloud data with image knowledge from various large-scale pre-trained 2D image models to solve the 3D point cloud few-shot classification problem.

\vspace*{8pt}
\subsection{Ablation Study}

\phy{To demonstrate the efficacy of different modules in our proposed scheme, we carry out ablative experiments with three backbones mentioned in the Subsection~\ref{Experimental Settings}.} 
Table \ref{tab:ablation_study} \bp{shows} the impact of different modules on MvNet for 16-shot classification results, including the Multi-view Attention Fusion, Conv Fusion, and the number of views. From the experimental results in \bp{this table,} \bp{we can see that} increasing the number of views can effectively improve the model's understanding of the point cloud object \bp{since oAcc gradually increases with more views.} 
At the same time, if the features obtained from the interaction between multiple views of the point cloud object are used as additional information for each view, and \bp{such} additional information is fully integrated with the original features through the synergy of the Attention Fusion module and the Conv Fusion module, the model's understanding of the point cloud object can be greatly improved. 
As shown in row 2 and 3 of Table~\ref{tab:ablation_study} respectively, using Attention Fusion to interact features from multiple views and adding the updated features directly to the original features, the oAcc is improved from $86.10\%$ to $87.84\%$,  $87.76\%$ to $88.12\%$, \bp{ and $86.42\%$ to $89.70\%$ for ResNet-18, ResNet-50, ConvNeXt-Large} respectively. 
\phy{By comparing row 3 and 5, we can also find that the use of Conv Fusion instead of direct summation results in a complete fusion of the multi-view features and therefore better model performance.} 


\vspace*{8pt}
\subsection{Visualization of Multi-view Vision Prompt}

\begin{figure}[tb]
\begin{center}
\vspace{6pt}
\includegraphics[width=0.70\linewidth]{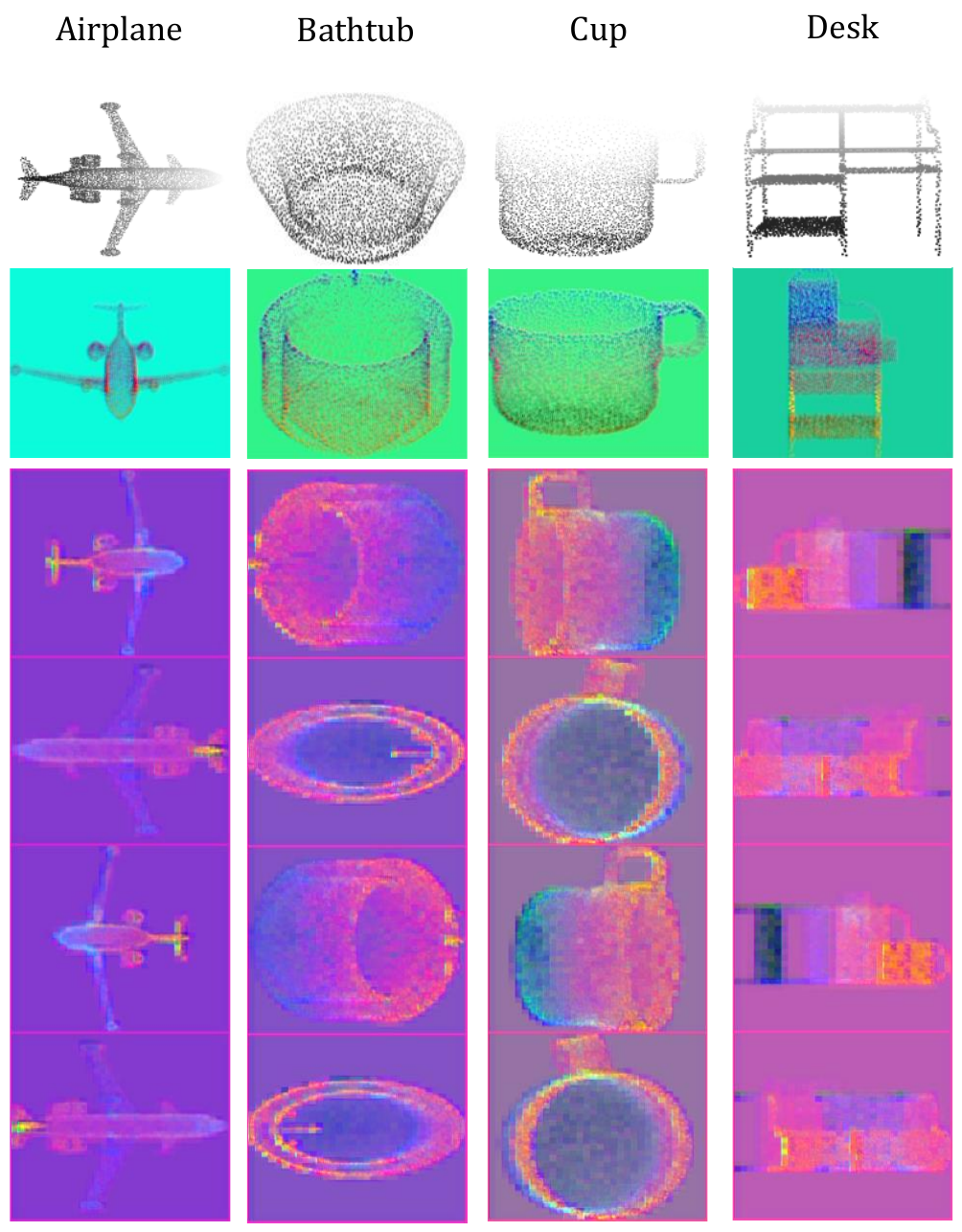}
\end{center}
\vspace{-0.1cm}
   \caption{\textbf{Images produced by single-view vision prompt and our Multi-view Vision Prompt Module.} We show three visualizations based on ModelNet: the original point cloud (top row), the visualization images from the single-view prompt (second row), and the visualization images from the multi-view prompt (rows three to six). As shown in the diagram, the multi-view vision prompts have richer, more nuanced color details.}
\label{fig: multi-view vision prompt}
\vspace{-0.35cm}
\end{figure}

\bp{To show the effects of our multi-view vision prompt, we illustrate some typical 3D point objects' visualization in Figure~\ref{fig: multi-view vision prompt}. The first row demonstrates some objects such as an airplane, bathtub, cup, and desk, and the second row contains one single view projection \phy{vision prompt}, while our multi-view prompt images are visualized from row $3$ - $6$.} \bp{As shown in Figure~\ref{fig: multi-view vision prompt}, the multi-view vision prompt can provide more rich information about the object, thus effectively boosting the feature representation capacity for 3D point object classification.} 

During the learning process of the Multi-view Vision Prompt module, it uses the multi-view vision prompts generated from the point clouds to learn a generic feature embedding for each type of point cloud object, using the multi-view features of the point cloud object and the guidance knowledge fed back from the large 2D pre-trained image model. A generic feature embedding \bp{can thus be obtained}  to distinguish which class the point cloud object belongs to. This process is similar to how humans recognize real-world objects in that we often group objects with similar characteristics together, e.g., the appearance of an aircraft consists of a pair of wings, a tail, and a set of engines. This idea is supported by the multi-view visualization in Figure~\ref{fig: multi-view vision prompt}. \bp{For instance,} looking at the visualization of the airplane in the first column of the diagram, the single view of the airplane in the second column has less color detail, and only the edges are clearly visible. In contrast, the multi-view projection shows more color detail, and we can see that the same parts of the airplane have similar colors. For example, the multi-view feature embedding of the tail of an airplane will appear distinctly yellow, while the head of the airplane will appear blue, and other similar areas will have the same feature visualization. It shows that the Multi-View Vision Prompt learns a more sophisticated feature embedding for different regions of a point cloud object
and thus has a more comprehensive understanding of the object.

\vspace*{8pt}
\section{Conclusion}

\bp{In this paper, to answer the question of whether a 2D pre-trained model can enhance 3D point cloud classification in the context of few-shot learning, we have proposed a novel multi-view vision prompt fusion network. Such a goal is achieved by projecting 3D points to multiple views, followed by an attention fusion module based on self-attention mechanism and feature fusion. The whole multi-view vision prompt fusion module is then taken as prompt for the pre-trained image model, much boosting the feature representation ability. Comprehensive comparison experiments and ablation studies validate the encouraging performance of the proposed scheme with different few-shots.} However, the large memory consumption is an important issue in the optimization process of the model, which will be our future research direction.

{\small
\bibliographystyle{unsrt}
\bibliography{egbib}

\begin{thebibliography}{10}

\bibitem{YukeZhu2017TargetdrivenVN}
Yuke Zhu, Roozbeh Mottaghi, Eric Kolve, Joseph~J. Lim, Abhinav Gupta,
  Li~Fei-Fei, and Ali Farhadi.
\newblock Target-driven visual navigation in indoor scenes using deep
  reinforcement learning.
\newblock {\em Cornell University - arXiv}, 2017.

\bibitem{YaodongCui2020DeepLF}
Yaodong Cui, Ren Chen, Wenbo Chu, Long Chen, Daxin Tian, and Dongpu Cao.
\newblock Deep learning for image and point cloud fusion in autonomous driving:
  A review.
\newblock {\em IEEE Transactions on Intelligent Transportation Systems}, 2020.

\bibitem{MingLiang2018DeepCF}
Ming Liang, Bin Yang, Shenlong Wang, and Raquel Urtasun.
\newblock Deep continuous fusion for multi-sensor 3d object detection.
\newblock {\em Cornell University - arXiv}, 2018.

\bibitem{WeijingShi2020PointGNNGN}
Weijing Shi, Ragunathan, and Rajkumar.
\newblock Point-gnn: Graph neural network for 3d object detection in a point
  cloud.
\newblock {\em arXiv: Computer Vision and Pattern Recognition}, 2020.

\bibitem{YuZhong2009IntrinsicSS}
Yu~Zhong.
\newblock Intrinsic shape signatures: A shape descriptor for 3d object
  recognition.
\newblock {\em International Conference on Computer Vision}, 2009.

\bibitem{RaduBogdanRusu2008AligningPC}
Radu~Bogdan Rusu, Nico Blodow, Zoltan-Csaba Marton, and Michael Beetz.
\newblock Aligning point cloud views using persistent feature histograms.
\newblock {\em Intelligent Robots and Systems}, 2008.

\bibitem{FedericoTombari2010UniqueSC}
Federico Tombari, Samuele Salti, and Luigi~Di Stefano.
\newblock Unique shape context for 3d data description.
\newblock 2010.

\bibitem{AlbertoGarciaGarcia2016PointNetA3}
Alberto Garcia-Garcia, Francisco Gomez-Donoso, Jose Garcia-Rodriguez, Sergio
  Orts-Escolano, Miguel Cazorla, and Jorge Azorin-Lopez.
\newblock Pointnet: A 3d convolutional neural network for real-time object
  class recognition.
\newblock {\em International Joint Conference on Neural Network}, 2016.

\bibitem{CharlesRQi2017PointNetDH}
Charles~R. Qi, Li~Yi, Hao Su, and Leonidas~J. Guibas.
\newblock Pointnet++: Deep hierarchical feature learning on point sets in a
  metric space.
\newblock {\em arXiv: Computer Vision and Pattern Recognition}, 2017.

\bibitem{AnhVietPhan2018DGCNNAC}
Anh~Viet Phan, Minh-Le Nguyen, Yen Lam~Hoang Nguyen, and Lam~Thu Bui.
\newblock Dgcnn: A convolutional neural network over large-scale labeled
  graphs.
\newblock {\em Neural Networks}, 2018.

\bibitem{YongbinLiao2023PointCI}
Yongbin Liao, Hongyuan Zhu, Yanggang Zhang, Chuangguan Ye, Tao Chen, and
  Jiayuan Fan.
\newblock Point cloud instance segmentation with semi-supervised bounding-box
  mining.
\newblock 2023.

\bibitem{SijinChen2023EndtoEnd3D}
Sijin Chen, Hongyuan Zhu, Xin Chen, Yinjie Lei, Tao Chen, and Gang YU.
\newblock End-to-end 3d dense captioning with vote2cap-detr.
\newblock 2023.

\bibitem{YulanGuo2019DeepLF}
Yulan Guo, Hanyun Wang, Qingyong Hu, Hao Liu, Li~Liu, and Mohammed Bennamoun.
\newblock Deep learning for 3d point clouds: A survey.
\newblock {\em Cornell University - arXiv}, 2019.

\bibitem{HangSu2015MultiviewCN}
Hang Su, Subhransu Maji, Evangelos Kalogerakis, and Erik Learned-Miller.
\newblock Multi-view convolutional neural networks for 3d shape recognition.
\newblock {\em arXiv: Computer Vision and Pattern Recognition}, 2015.

\bibitem{TanYu2018MultiviewHB}
Tan Yu, Jingjing Meng, and Junsong Yuan.
\newblock Multi-view harmonized bilinear network for 3d object recognition.
\newblock {\em Computer Vision and Pattern Recognition}, 2018.

\bibitem{YuxingXie2019ARO}
Yuxing Xie, Jiaojiao Tian, and Xiao~Xiang Zhu.
\newblock A review of point cloud semantic segmentation.
\newblock 2019.

\bibitem{GusiTe2018RGCNNRG}
Gusi Te, Wei Hu, Amin Zheng, and Zongming Guo.
\newblock Rgcnn: Regularized graph cnn for point cloud segmentation.
\newblock {\em ACM Multimedia}, 2018.

\bibitem{JiaDeng2009ImageNetAL}
Jia Deng, Wei Dong, Richard Socher, Li-Jia Li, Kai Li, and Li~Fei-Fei.
\newblock Imagenet: A large-scale hierarchical image database.
\newblock {\em Computer Vision and Pattern Recognition}, 2009.

\bibitem{TsungYiLin2014MicrosoftCC}
Tsung-Yi Lin, Michael Maire, Serge Belongie, James Hays, Pietro Perona, Deva
  Ramanan, Piotr Doll{\'a}r, and C.~Lawrence Zitnick.
\newblock Microsoft coco: Common objects in context.
\newblock {\em Lecture Notes in Computer Science}, 2014.

\bibitem{MarkEveringham2010ThePV}
Mark Everingham, Luc~Van Gool, Christopher Williams, John Winn, and Andrew
  Zisserman.
\newblock The pascal visual object classes (voc) challenge.
\newblock {\em International Journal of Computer Vision}, 2010.

\bibitem{KaimingHe2015DeepRL}
Kaiming He, Xiangyu Zhang, Shaoqing Ren, and Jian Sun.
\newblock Deep residual learning for image recognition.
\newblock {\em Cornell University - arXiv}, 2015.

\bibitem{ZhuangLiu2023ACF}
Zhuang Liu, Hanzi Mao, Chao-Yuan Wu, Christoph Feichtenhofer, Trevor Darrell,
  Saining Xie, and A~Facebook.
\newblock A convnet for the 2020s.
\newblock 2023.

\bibitem{AlexeyDosovitskiy2020AnII}
Alexey Dosovitskiy, Lucas Beyer, Alexander Kolesnikov, Dirk Weissenborn,
  Xiaohua Zhai, Thomas Unterthiner, Mostafa Dehghani, Matthias Minderer, Georg
  Heigold, Sylvain Gelly, Jakob Uszkoreit, and Neil Houlsby.
\newblock An image is worth 16x16 words: Transformers for image recognition at
  scale.
\newblock {\em arXiv: Computer Vision and Pattern Recognition}, 2020.

\bibitem{ZhirongWu20143DSA}
Zhirong Wu, Shuran Song, Aditya Khosla, Fisher Yu, Linguang Zhang, Xiaoou Tang,
  and Jianxiong Xiao.
\newblock 3d shapenets: A deep representation for volumetric shapes.
\newblock {\em Cornell University - arXiv}, 2014.

\bibitem{MikaelaAngelinaUy2019RevisitingPC}
Mikaela~Angelina Uy, Quang-Hieu Pham, Binh-Son Hua, Duc~Thanh Nguyen, and
  Sai-Kit Yeung.
\newblock Revisiting point cloud classification: A new benchmark dataset and
  classification model on real-world data.
\newblock {\em arXiv: Computer Vision and Pattern Recognition}, 2019.

\bibitem{AngelXChang2015ShapeNetAI}
Angel~X. Chang, Thomas Funkhouser, Leonidas~J. Guibas, Pat Hanrahan, Qixing
  Huang, Zimo Li, Silvio Savarese, Manolis Savva, Shuran Song, Hao Su,
  Jianxiong Xiao, Li~Yi, and Fisher Yu.
\newblock Shapenet: An information-rich 3d model repository.
\newblock {\em arXiv: Graphics}, 2015.

\bibitem{ye2022makes}
Chuangguan Ye, Hongyuan Zhu, Yongbin Liao, Yanggang Zhang, Tao Chen, and
  Jiayuan Fan.
\newblock What makes for effective few-shot point cloud classification?
\newblock In {\em Proceedings of the IEEE/CVF winter conference on applications
  of computer vision}, pages 1829--1838, 2022.

\bibitem{ye2023closer}
Chuangguan Ye, Hongyuan Zhu, Bo~Zhang, and Tao Chen.
\newblock A closer look at few-shot 3d point cloud classification.
\newblock {\em International Journal of Computer Vision}, 131(3):772--795,
  2023.

\bibitem{RenruiZhang2023PointCLIPPC}
Renrui Zhang, Ziyu Guo, Wei Zhang, Kunchang Li, Xupeng Miao, Bin Cui, Yu~Qiao,
  Peng Gao, and Hongsheng Li.
\newblock Pointclip: Point cloud understanding by clip.
\newblock 2023.

\bibitem{XiangyangZhu2022PointCLIPVA}
Xiangyang Zhu, Renrui Zhang, Bowei He, Ziyao Zeng, Shanghang Zhang, and Peng
  Gao.
\newblock Pointclip v2: Adapting clip for powerful 3d open-world learning.
\newblock 2022.

\bibitem{XiangLisaLi2021PrefixTuningOC}
Xiang~Lisa Li and Percy Liang.
\newblock Prefix-tuning: Optimizing continuous prompts for generation.
\newblock {\em Cornell University - arXiv}, 2021.

\bibitem{BrianLester2021ThePO}
Brian Lester, Rami Al-Rfou, and Noah Constant.
\newblock The power of scale for parameter-efficient prompt tuning.
\newblock 2021.

\bibitem{XiaoLiu2021GPTUT}
Xiao Liu, Yanan Zheng, Zhengxiao Du, Ming Ding, Yujie Qian, Zhilin Yang, and
  Jie Tang.
\newblock Gpt understands, too.
\newblock {\em arXiv: Computation and Language}, 2021.

\bibitem{yuan2023ad}
Jiakang Yuan, Bo~Zhang, Xiangchao Yan, Tao Chen, Botian Shi, Yikang Li, and
  Yu~Qiao.
\newblock Ad-pt: Autonomous driving pre-training with large-scale point cloud
  dataset.
\newblock {\em arXiv preprint arXiv:2306.00612}, 2023.

\bibitem{zhang2023uni3d}
Bo~Zhang, Jiakang Yuan, Botian Shi, Tao Chen, Yikang Li, and Yu~Qiao.
\newblock Uni3d: A unified baseline for multi-dataset 3d object detection.
\newblock In {\em Proceedings of the IEEE/CVF Conference on Computer Vision and
  Pattern Recognition}, pages 9253--9262, 2023.

\bibitem{AmirRoshanZamir2018TaskonomyDT}
Amir~Roshan Zamir, Alexander Sax, William~B. Shen, Leonidas~J. Guibas, Jitendra
  Malik, and Silvio Savarese.
\newblock Taskonomy: Disentangling task transfer learning.
\newblock 2018.

\bibitem{KarenSimonyan2015VeryDC}
Karen Simonyan and Andrew Zisserman.
\newblock Very deep convolutional networks for large-scale image recognition.
\newblock {\em International Conference on Learning Representations}, 2015.

\bibitem{PengYe2022StimulativeTO}
Peng Ye, Shengji Tang, Baopu Li, Tao Chen, and Wanli Ouyang.
\newblock Stimulative training of residual networks: A social psychology
  perspective of loafing.
\newblock 2022.

\bibitem{GaoHuang2016DenselyCC}
Gao Huang, Zhuang Liu, Laurens van~der Maaten, and Kilian~Q. Weinberger.
\newblock Densely connected convolutional networks.
\newblock {\em arXiv: Computer Vision and Pattern Recognition}, 2016.

\bibitem{HangboBao2021BEiTBP}
Hangbo Bao, Li~Dong, and Furu Wei.
\newblock Beit: Bert pre-training of image transformers.
\newblock {\em Cornell University - arXiv}, 2021.

\bibitem{KaimingHe2021MaskedAA}
Kaiming He, Xinlei Chen, Saining Xie, Yanghao Li, Piotr Doll{\'a}r, and Ross
  Girshick.
\newblock Masked autoencoders are scalable vision learners.
\newblock {\em arXiv: Computer Vision and Pattern Recognition}, 2021.

\bibitem{KaimingHe2023MomentumCF}
Kaiming He, Haoqi Fan, Yuxin Wu, Saining Xie, and Ross Girshick.
\newblock Momentum contrast for unsupervised visual representation learning.
\newblock 2023.

\bibitem{AnttiTarvainen2017MeanTA}
Antti Tarvainen and Harri Valpola.
\newblock Mean teachers are better role models: Weight-averaged consistency
  targets improve semi-supervised deep learning results.
\newblock {\em International Conference on Learning Representations}, 2017.

\bibitem{DavidBerthelot2019MixMatchAH}
David Berthelot, Nicholas Carlini, Ian Goodfellow, Nicolas Papernot, Avital
  Oliver, and Colin Raffel.
\newblock Mixmatch: A holistic approach to semi-supervised learning.
\newblock {\em arXiv: Learning}, 2019.

\bibitem{QizheXie2020SelfTrainingWN}
Qizhe Xie, Minh-Thang Luong, Eduard Hovy, and Quoc~V. Le.
\newblock Self-training with noisy student improves imagenet classification.
\newblock {\em Computer Vision and Pattern Recognition}, 2020.

\bibitem{SainingXie2020PointContrastUP}
Saining Xie, Jiatao Gu, Demi Guo, Charles~R. Qi, Leonidas~J. Guibas, and
  Or~Litany.
\newblock Pointcontrast: Unsupervised pre-training for 3d point cloud
  understanding.
\newblock {\em Cornell University - arXiv}, 2020.

\bibitem{XuminYu2023PointBERTP3}
Xumin Yu, Lulu Tang, Yongming Rao, Tiejun Huang, Jie Zhou, and Jiwen Lu.
\newblock Point-bert: Pre-training 3d point cloud transformers with masked
  point modeling.
\newblock 2023.

\bibitem{pang2022masked}
Yatian Pang, Wenxiao Wang, Francis E.~H. Tay, Wei Liu, Yonghong Tian, and
  Li~Yuan.
\newblock Masked autoencoders for point cloud self-supervised learning, 2022.

\bibitem{AngelaDai2017ScanNetR3}
Angela Dai, Angel~X. Chang, Manolis Savva, Maciej Halber, Thomas Funkhouser,
  and Matthias NieBner.
\newblock Scannet: Richly-annotated 3d reconstructions of indoor scenes.
\newblock {\em Computer Vision and Pattern Recognition}, 2017.

\bibitem{XiaoLiu2023PTuningVP}
Xiao Liu, Kaixuan Ji, Yicheng Fu, Zhengxiao Du, Zhilin Yang, and Jie Tang.
\newblock P-tuning v2: Prompt tuning can be comparable to fine-tuning
  universally across scales and tasks.
\newblock 2023.

\bibitem{MenglinJia2022VisualPT}
Menglin Jia, Luming Tang, Bor-Chun Chen, Claire Cardie, Serge Belongie, Bharath
  Hariharan, and Ser-Nam Lim.
\newblock Visual prompt tuning.
\newblock 2022.

\bibitem{wang2022p2p}
Ziyi Wang, Xumin Yu, Yongming Rao, Jie Zhou, and Jiwen Lu.
\newblock P2p: Tuning pre-trained image models for point cloud analysis with
  point-to-pixel prompting.
\newblock {\em arXiv preprint arXiv:2208.02812}, 2022.

\bibitem{yuan2023bi3d}
Jiakang Yuan, Bo~Zhang, Xiangchao Yan, Tao Chen, Botian Shi, Yikang Li, and
  Yu~Qiao.
\newblock Bi3d: Bi-domain active learning for cross-domain 3d object detection.
\newblock In {\em Proceedings of the IEEE/CVF Conference on Computer Vision and
  Pattern Recognition}, pages 15599--15608, 2023.

\bibitem{AlecRadford2021LearningTV}
Alec Radford, Jong~Wook Kim, Chris Hallacy, Aditya Ramesh, Gabriel Goh,
  Sandhini Agarwal, Girish Sastry, Amanda Askell, Pamela Mishkin, Jack Clark,
  Gretchen Krueger, and Ilya Sutskever.
\newblock Learning transferable visual models from natural language
  supervision.
\newblock {\em arXiv: Computer Vision and Pattern Recognition}, 2021.

\bibitem{AshishVaswani2017AttentionIA}
Ashish Vaswani, Noam Shazeer, Niki Parmar, Jakob Uszkoreit, Llion Jones,
  Aidan~N. Gomez, Lukasz Kaiser, and Illia Polosukhin.
\newblock Attention is all you need.
\newblock {\em Neural Information Processing Systems}, 2017.

\bibitem{AAMMuzahid2021CurveNetCM}
A.~A.~M. Muzahid, Wanggen Wan, Ferdous Sohel, Lianyao Wu, and Li~Hou.
\newblock Curvenet: Curvature-based multitask learning deep networks for 3d
  object recognition.
\newblock {\em IEEE/CAA Journal of Automatica Sinica}, 2021.

\bibitem{AnkitGoyal2021RevisitingPC}
Ankit Goyal, Hei Law, Bowei Liu, Alejandro Newell, and Jia Deng.
\newblock Revisiting point cloud shape classification with a simple and
  effective baseline.
\newblock {\em Cornell University - arXiv}, 2021.

\bibitem{DanHendrycks2016GaussianEL}
Dan Hendrycks and Kevin Gimpel.
\newblock Gaussian error linear units (gelus).
\newblock {\em Cornell University - arXiv}, 2016.

\bibitem{SergeyIoffe2015BatchNA}
Sergey Ioffe and Christian Szegedy.
\newblock Batch normalization: Accelerating deep network training by reducing
  internal covariate shift.
\newblock {\em arXiv: Learning}, 2015.

\bibitem{IlyaLoshchilov2017DecoupledWD}
Ilya Loshchilov and Frank Hutter.
\newblock Decoupled weight decay regularization.
\newblock {\em arXiv: Learning}, 2017.

\bibitem{IlyaLoshchilov2016SGDRSG}
Ilya Loshchilov and Frank Hutter.
\newblock Sgdr: Stochastic gradient descent with warm restarts.
\newblock {\em arXiv: Learning}, 2016.

\bibitem{GuotaiWang2019AleatoricUE}
Guotai Wang, Wenqi Li, Michael Aertsen, Jan Deprest, Sebastien Ourselin, and
  Tom Vercauteren.
\newblock Aleatoric uncertainty estimation with test-time augmentation for
  medical image segmentation with convolutional neural networks.
\newblock {\em Neurocomputing}, 2019.

\end{thebibliography}
}

\end{document}